\documentclass{article}
\usepackage[papersize={210mm,297mm}, total={160mm,250mm}]{geometry}

\usepackage[square,numbers]{natbib}
\usepackage[utf8]{inputenc} 
\usepackage[T1]{fontenc} 
\usepackage{hyperref} 
\usepackage{url} 
\usepackage{booktabs} 
\usepackage{amsfonts} 
\usepackage{nicefrac} 
\usepackage{microtype} 
\usepackage{lipsum}
\usepackage{fancyhdr} 
\usepackage{graphicx} 
\graphicspath{{media/}} 
\usepackage{xcolor}
\usepackage[para]{threeparttable}

\usepackage{amsmath}
\usepackage{nccmath}
\usepackage{tabularx}
\usepackage{multirow}
\usepackage{amssymb}
\usepackage{subfig}

\usepackage{soul}
\usepackage{color}
\sethlcolor{white}
\usepackage{listings}

\graphicspath{ {images/} }
\title{A Prescriptive Learning Analytics Framework: Beyond Predictive Modelling and onto Explainable AI with Prescriptive Analytics \hl{and ChatGPT}

}

\author{
  Teo Susnjak\footnote{\href{mailto:t.susnjak@massey.ac.nz}{Email: t.susnjak@massey.ac.nz}} \\
  School of Mathematical and Computational Sciences \\
  Massey University \\
  Auckland, New Zealand \\
}

\begin{document}
\maketitle

\begin{abstract}
A significant body of recent research in the field of Learning Analytics has focused on leveraging machine learning approaches for predicting at-risk students in order to initiate timely interventions and thereby elevate retention and completion rates. The overarching feature of the majority of these research studies has been on the science of prediction only. The component of predictive analytics concerned with interpreting the internals of the models and explaining their predictions for individual cases to stakeholders has largely been neglected. Additionally, works that attempt to employ data-driven prescriptive analytics to automatically generate evidence-based remedial advice for at-risk learners are in their infancy. 

eXplainable AI is a field that has recently emerged providing cutting-edge tools which support transparent predictive analytics and techniques for generating tailored advice for at-risk students. This study proposes a novel framework that unifies both transparent machine learning as well as techniques for enabling prescriptive analytics\hl{, while integrating the latest advances in large language models}. This work practically demonstrates the proposed framework using predictive models for identifying at-risk learners of programme non-completion. The study then further demonstrates how predictive modelling can be augmented with prescriptive analytics on two case studies in order to generate human-readable prescriptive feedback for those who are at risk\hl{ using ChatGPT}.
\end{abstract}



\maketitle

\section{Introduction}\label{}

Higher Education Institutions (HEI) are operating in an ever-increasingly competitive and dynamic environment. The recently accelerated transition to online learning by many institutions has been one of the key drivers of these emerging pressures. Within this context, student retention rates have also become a prominent issue at HEI, as well as concerns around student performance in general due to the prevalence of low grades \cite{namoun2020predicting}. Considerable research effort has been invested into technological means to address these issues and the Learning Analytics (LA) field, in general, has been an important vehicle for supporting these endeavours \cite{wong2020review}.

Within LA, it is widely appreciated that predictive analytics tools that identify at-risk students hold considerable potential to address these challenges at least in part, by providing the ability for timely interventions to be initiated with at-risk learners which can result in corrective measures being undertaken by them \cite{namoun2020predicting}. However, in their survey of LA applications, \citet{hernandez2022learning} conclude that LA technologies are generally not yet widely used in this sector despite the evident potential they offer to HEI. Indeed, \citet{jang2022practical} highlight that despite the clear opportunities offered by the predictive analytics technologies, the developed tools tend to persist only as research content. The authors posit that to genuinely integrate the predictive analytics technologies into educational contexts, barriers like educators’ general distrust in the tools, as well as the lack of interpretability and visual representation of information accompanying them, need to be overcome and are perceived as being some of the biggest obstacles. \citet{rets2021exploring} also point out that learners will engage with a LA tools only if they can understand how the LA system's outputs regarding them are generated.

Predictive models themselves have over time become more complex and as a result, they have assumed 'black-box' characteristics. Their complexity ensures that it is not possible to apprehend how these models arrive at their predictions, and crucially, what aspects of the learners’ behaviours are key determinants of their prognosticated outcomes. This absence of model \textit{interpretability} (how a model works) and \textit{explainability} of their specific predictions in turn negatively affects their utility, and ultimately results in distrust by the stakeholders \cite{baneres2021predictive}.

According to \citet{villagra2017improving}, for a prediction model to truly be useful, it ought to also do more than merely classify learners into risk categories. The authors argue that the models should in addition offer interpretative characteristics from which learners can gain insights into possible causes of their learning obstacles. Transparency of this kind facilitates the development of trust by all stakeholders in general and thus increases the prospects of adoption. The greater interpretability of the predictive models enhances the tool's capabilities to support the dispensing of effective guidance towards resolution and remediation to at-risk learners. In their recent systematic literature review of predictive LA studies, \citet{namoun2020predicting} also encouraged the pursuit of research into interpretability and explainability of the predictive models where the focus should more heavily focus on developing explanatory aspects of predictions rather than the development of models that merely forecast student outcomes. 

In the most ideal setting, the analytics technologies used in the educational contexts need to even go beyond interpretability and explainability properties, and should in fact embody \textit{prescriptive} analytics capabilities which leverage data-driven techniques to communicate to learners precisely what remedial actions are most likely to result in improved outcomes \cite{susnjak2022learning}. While descriptive analytics answers \textit{'what happened?'} and predictive analytics addresses \textit{'what will happen?'}, prescriptive analytics tackles \textit{'how to make it happen?'} \cite{frazzetto2019prescriptive}. The power of prescriptive analytics, therefore, lies in its ability to transform information into implementable decisions. \citet{liu2017going} highlight the importance of LA tools which lead directly to actionability. Arguably, therefore, the most beneficial and insight-rich form of analytics is found in the prescriptive data-driven outputs which generate the greatest intelligence and value \cite{lepenioti2020prescriptive}.

Prescriptive analytics frequently uses machine learning in order to suggest at the existence of possible causal relationships within the features describing learners, and consequently, recommendations can be constructed from these outputs which can be used as advice concerning which behavioural adjustments are likely to result in more desirable outcomes. By offering tailored and evidence-based recommendations that venture beyond generic advice, and are instead customised to each learner with specific and measurable goals, more effective advice can be provided to support interventions, which \citet{wong2020review} confirm in their recent review of learning analytics intervention studies. Thereby, the prospects of positively affecting both retention rates and student performances are consequently improved. Increasingly studies are emerging that have highlighted the importance of LA tools which provide insights to learners with a prescriptive component that are in the form of recommendations for guiding learners \cite{valle2021predict} and this is an emerging research frontier.

\hl{While tailored learner recommendations in the form of everyday natural language can be automatically constructed from the numerical outputs of prescriptive analytics using natural language processing (NLP) tools, this is not straightforward. NLP toolkits and significant development effort is required in order to develop reliable software which produces grammatically correct natural language outputs based on variable numerical outputs. Given the recent advances in transformer-based large language models (LLMs) such as GPT-series and ChatGPT specifically, these technologies provide the capability to bridge this non-trivial challenge. }


\subsection{eXplainable AI hl{ and LLMs}}

With the increased embedding of complex predictive models into contexts that were previously dominated by human decision-making, the need has arisen for predictive models to display a greater degree of transparency behind their mechanisms of reasoning. This is the case not least for establishing trust in them, but also from the perspective of compliance. It is increasingly becoming a legal requirement across international jurisdictions\footnote{GDPR \cite{regulation2016regulation} is one example.} that the mechanisms behind any automated decisions affecting humans be clearly explained to those affected by them. \citet{wachter2017counterfactual} also point out the necessity of being able to contest decisions made by automated systems by those concerned and to also have the opportunity to be informed with respect to the current decision-making model as to what it would need to see change in their data inputs to produce an alternative or a more desirable decision.

It is for these reasons that a relatively new research field of eXplainable Artificial Intelligence (XAI) (sometimes referred to as Interpretable Machine Learning) has emerged. Some of the key aspirations of this field are focused on developing techniques that address the high-level interpretability of opaque predictive models as well as on devising tools that enable them to interrogate and extract explanations of how the models arrive at given conclusions \cite{molnar2020interpretable}. Meanwhile, there currently exists a gathering the research interest exploring prescriptive analytics where methods are being developed that leverage data-driven approaches to generate evidence-based actionable insights \cite{lepenioti2020prescriptive}.

From a technical point of view, model interpretability tends to pertain to the tasks of making sense of a model's internals generated post-training by a machine learning algorithm. It usually concerns a level of clarity into the mechanics of a model at a \textit{global-level}, while explainability of models refers to extracting the reasoning behind the model's prediction for a specific learner in this context, which is referred to as \textit{local-level} explainability. Both perspectives on model behaviour are important and serve multiple overlapping purposes. Model interpretability affords an institution the ability to communicate to all concerned stakeholders how a predictive model works in general terms using broad brush strokes. While local-level explainability of models enables the validation of specific predictions to take place by student support teams before interventions are initiated, it also enables clear responses to be given to affected students as to exactly how and why they have been identified as being at-risk. However, clues can also be gleaned from the local-level explanations into possible remedial actions that can be suggested to the learners.

Tools and techniques supporting the goals of model transparency are reaching maturity. SHAP \cite{lundberg2017unified} is a visualisation technique that is currently recognised as being state-of-the-art in the field of XAI \cite{gramegna2021shap} for realising both global and local-level model transparency. The Anchors \cite{ribeiro2018anchors} technique has also been developed recently as a tool that imparts a high degree of local-level explainability. The benefit of Anchors is that it produces human-readable rule-based models which are succinct approximations of the behaviour of the underlying complex models. However, a more advanced approach using Counterfactuals \cite{wachter2017counterfactual} enables the analytics processes to surpass predictive capabilities, and enter into the prescriptive. This technique provides precise suggestions to the learners regarding the smallest set of adjustments they need to make in their learning behaviour for an alternative prediction to be generated. 
This study demonstrates the applicability of all three of these technologies for different steps of the proposed prescriptive analytics framework. 

\hl{Meanwhile, there have been significant advances in the field of AI and LLMs based on transformers which are now approaching human-level capabilities for text generation and discourse. Driven largely by their ability to capture long-range dependencies and contextual relationships in text through the use of self-attention mechanisms, models such as Google's BERT \mbox{\cite{devlin2018bert}} and the latest GPT-series from OpenAI, have achieved state-of-the-art results in various natural language processing tasks, including text generation \mbox{\cite{Brown2020}}. These advances continue and have been demonstrated by OpenAI's latest ChatGPT model, which has the potential to translate complex analytics outputs into human-readable and actionable natural language for learners and student advisors.}

\subsection{Research Contribution}

This study presents a novel prescriptive analytics framework for supporting LA aims of identifying and initiating both timely and effective interventions with at-risk students. The proposed framework demonstrates how both predictive and prescriptive analytics can be more fully leveraged than has previously been done. This study illustrates how effective models predicting qualification completion outcomes can be developed using machine learning and made transparent at both global and local levels to meet the needs of all stakeholders. Furthermore, this work goes on to illustrate through case study examples how prescriptive analytics tools can subsequently be utilised to automatically generate specific prescriptive feedback \hl{which is translated into natural language with the assistance of ChatGPT}, providing suggestions that are evidence-based as well as actionable.

\section{Background}\label{}

In the last two years, there have been numerous systematic literature reviews (SLR) in the field of LA and Educational Data Mining (EDM) centring on predictive themes, casting multiple perspectives onto the state of this field, and revealing where the focus of the research efforts. Indeed, these works have identified current gaps, emerging trends, and future aspirations which are brought out in the following sections. 

\subsection{Recent Survey Findings}
An in-depth SLR was conducted by \citet{namoun2020predicting}, covering a total of 62 relevant studies within predictive LA, in which they focused on three areas: (1) ways in which academic performance was measured using learning outcomes and subsequently predicted, (2) the types of algorithms used to forecast student learning outcomes, and (3) which features are most impactful for predicting student outcomes. The authors found that 90\% of the studies predicted course-level outcomes, while only three studies considered predicting programme/qualification-level outcomes. As part of their findings, they urged the research community to conduct further work in predictive modelling, and more specifically, to do so at the programme level which they described presently to be in its infancy. Meanwhile, the authors called researchers to rise above simply predicting outcomes, and to also incorporate model interpretability and explainability into their studies. 

\citet{albreiki2021systematic} likewise conducted an SLR consisting of 78 studies of relevant EDM literature from 2009 to 2021 concerning predicting at-risk learners of non-completion. The review indicated that only a handful of studies proposed means for generating remedial solutions consisting of feedback that learners and educators can use to address the underlying obstacles. The authors noted that future research will place greater emphasis on devising machine learning methods to predict students’ performance in general and will also augment this with automatically generated remedial actions to assist learners as early as possible.

Another up-to-date survey by \citet{xiao2022survey} examined almost 80 studies using EDM to predict students’ performance and provided insights in line with previous studies. The authors note that one of the deficiencies in this field is that very few studies have attended to explore model interpretability and have thus neglected explanations of the mechanics of predictions and the role that different features play in the predicted outputs. Indeed, the authors stress that the use of model interpretability tools ought to be one of the chief pursuits and the direction of future research in this field of study.

There were several other recently conducted reviews into LA and EDM studies. Interestingly, these did not address the issue of model interpretability and explainability, nor was there coverage of the use of prescriptive analytics tools indicating that these approaches are not yet in wide use. \citet{fahd2021application} carried out a broad meta-analysis of literature of 89 studies from 2010 to 2020. Their survey analysed the application of machine learning approaches in predicting student academic performance. Their primary focus was on considering what types of models were being used in research at HEI, and their identification of trends centred around specific matters of how to achieve better predictive accuracies using machine learning. 
Similarly, \citet{batool2022educational} conducted a LA survey of some 260 studies over the last 20 years on the topic of student outcome prediction. Similar to previous reviews, they focused on highlighting the most effective features for this task and the types of algorithms and techniques used, as well as data mining tools that are most frequently applied. However, the important issues of model interpretability and prescriptive analytics were not raised. 

Likewise, \citet{shafiq2022student} conducted an SLR covering 100 papers. Their focus was again on predictive analytics, looking mostly at what was being predicted, what kind of data was used, which sets of features were effective or otherwise, and what types of algorithms were explored. The focus was essentially on ways to enhance the accuracy of predictive models but there was no coverage of techniques that extend beyond mere predictions
Instead, the recommendations emphasised that more ensemble-based and clustering methods should be explored to predict the performance of students and enhance the prediction accuracy. In the same vein, \citet{tjandra2022student}  conducted a comprehensive review of recent studies based on student performance prediction. The review considered features and algorithms used, accuracies attained, as well as commonly used tools. The authors concluded that there is still limited use of personal characteristics data such as psychological and social/behavioural features for developing student performance predictions and that future research should focus on including these to address dropout rates. Finally, \citet{hernandez2022learning} investigated the practices of 16 HEIs that have deployed LA projects. The authors found that they have mostly used LA technologies for student retention. These tools largely supported strategies for identifying at-risk students through predictive analytics, which served as a springboard for initiating various types of interventions.

A vast majority of the predictive models identifying at-risk students have focused on course-level outcomes rather than on programme-level completions \cite{namoun2020predicting}. Since course-level predictive models have tended to only be deployed across subsets of all courses on offer, the likelihood of at-risk students evading detection is therefore high. Thus, there is high utility in pursuing programme-level completion predictions. Though prior works in the prediction of programme completions are very rare, those that exist have generally followed an outcomes-based approach, which breaks down all constituent parts of programme-level outcome requirements into course outcomes first, and subsequently proceeds to map them all to programmes. These types of bottom-up predictive models then operate on the level of course-level outcomes and these predictions are combined to generate programme-level outcome predictions.
Examples of this approach are \cite{dandin2018attainment}, as well as \cite{bhatia2017automated} and \cite{bindra2017outcome} who likewise developed programme-completion prediction models and claimed to achieve accuracies in the upper ranges between 90\% and 95\% using various techniques from the WEKA data mining toolkit. More recently, \citet{Gupta2021} followed a similar approach in first determining course-level outcome predictions and mapping these to programmes, before determining overall outcomes. Such systems embody a great deal of complexity and appear to have been mostly research prototypes since productionisation and long-term maintenance of these approaches in a dynamic setting are questionable with respect to sustainability.  

\subsection{Model Interpretability in LA}

Model interpretability is not only important for building trust, meeting compliance and extracting maximal value from predictive systems, but it is also vital for developing accurate predictive models. Feature engineering and selection tasks are more critical to the success of machine learning models than the choice of algorithm or the size of the datasets used \cite{domingos2012few}. While a wide range of features for predicting student performance have been used in literature, there is still a lack of clarity on which specific features are most effective and how they interact to influence the attainment of course and programme outcomes \cite{namoun2020predicting}. Therefore, tools and approaches that specifically illuminate the influence and the behaviour of the models in terms of the underlying features, are important. In one of the earliest studies into methods of illuminating the predictions of black-box models for learners, \citet{villagra2017improving} developed a set of proprietary graphical tools to exploit the output information and provide a meaningful guide to both learners and instructors. 
\citet{dass2021predicting} develop dropout prediction models for students in a MOOC course and did strive to gain a deeper understanding of the effectiveness of the various features to support accurate predictions, as well as to predict the point in time when the students were likely to drop out.
Similarly, \citet{jang2022practical} demonstrate the usage of XAI techniques to assist in interpreting the classification results of the models which were designed to identify at-risk students. The study paid particular attention to selecting features that were relevant to stakeholders.

\subsection{Prescriptive LA Systems}
\citet{jenhani2016course} developed one of the earliest automated remedial prescriptive systems which leveraged machine learning. The system was designed in such a way that a separate classification model was trained on a remedial action dataset describing historical data based on experts’ and instructors’ actions to improve the low learning outcomes. The types of predictive outputs that the system made as suggestions for at-risk students were: revise a concept, attempt extra quizzes, solve specific practice examples, take extra assignments, etc. All these prescriptions were at a generic level. \citet{elhassan2018remedial} extended this work further to enable the system to recommend a set of remedial actions to address specific shortcomings rather than a single action as previously. 

\citet{albreiki2021customized} developed predictive models for at-risk students which they combined with a customised system that enabled the instructor to set various thresholds and weights alongside the predictive models. Based on the configuration and the outputs of the predictive models, the system would then select from a predefined list of remedial actions the most suitable one. 

Most recently, \citet{susnjak2022learning} proposed a student-oriented LA dashboard which apart from using descriptive and predictive analytics, also demonstrated a prototype of how to integrate interpretability and explainability aspects of the predictive models, as well as the automatic generation of feedback based on prescriptive analytics. 


\subsection{Summary and Research Aims}

From this literature review, several conclusions can be drawn. The literature indicates that the use of predictive analytics within the LA field is widespread and on the rise. The literature points to an acute gap in the current research concerning the use of predictive analytics tools which enable interpretability and explainability of the models. The bulk of the research is still largely focused on devising means of more effectively predicting various outcomes. There is a general absence of XAI tools in usage. These tools have only just started to emerge in LA research even though these tools are not esoteric. Much less is there any evidence that there exists the use of data-driven prescriptive analytics which provide automated and tangible remedial advice to at-risk students. Thus the conversion of predictions into actionable insights is broadly missing from the literature at this point. 

There is therefore a need and an emerging requirement to begin to implement predictive analytics systems with responsible and accountable characteristics. These LA systems ought to have embedded transparency that is enabled by an expressive set of technologies, while venturing beyond the predictive realm and ideally being augmented by prescriptive tools that assist learners in addressing their challenges. Unless an effective predictive system can be matched with the reasoning behind its decisions surrounding the underlying variables, and suggestions about concrete pathways for moving forward, these systems will only ever have constrained capabilities and limited uptake.

To that end, this work proposes a prescriptive learning analytics framework that attempts to address the outlined gaps. This work defines a step-by-step process for building predictive models and highlights ways for enhancing them with various levels of interpretability, while ultimately showing how prescriptive analytics can complement the entire process. This work demonstrates the proposed framework using programme/qualification completion predictive models and maps each step of the framework with specific technologies while showing the practicality of the framework with use case scenarios.
 
\section{Proposed Framework}\label{}
\begin{figure}[hbt]
	\centering
		\includegraphics[scale=0.4]{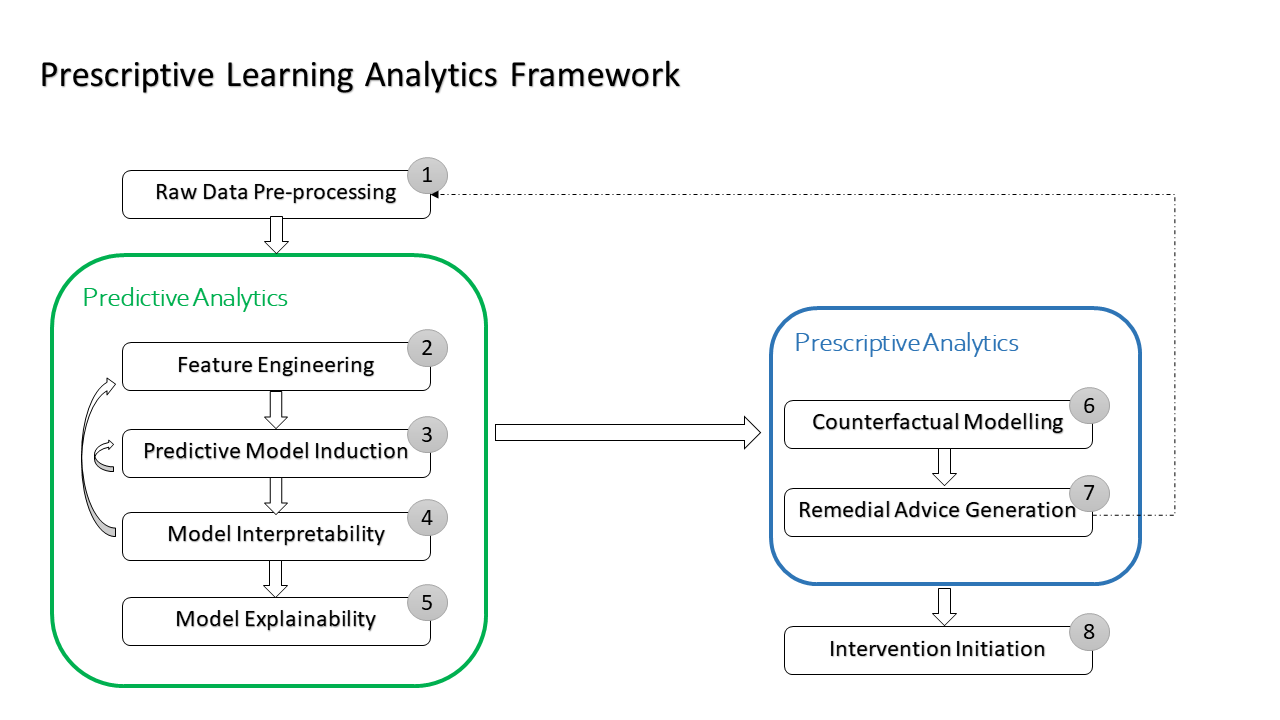}
	  \caption{ The Prescriptive Learning Analytics Framework (PLAF) highlights each step in the process. }\label{claf}
\end{figure}

The proposed Prescriptive Learning Analytics Framework (PLAF) is outlined here and depicted in Figure \ref{claf}. The figure portrays the two key components, namely predictive and prescriptive phases, as well as the process flow of the various steps comprising the framework. The framework assumes that the data identification and acquisition steps have already been completed. Each step in the framework is discussed in turn.  

The initial Step (1) assumes that the raw data has been cleaned and preprocessed to support subsequent analyses. Here, all relevant exploratory data analyses are performed, and an investigation into the reliability of the data. This includes investigating the prevalence of missing values, together with methods to impute them if necessary. Additionally, the ability of the existing data to support deeper analyses of the subsequent phases is conducted here.

\subsubsection*{Predictive Analytics Phase}

Predictive analytics begins with the feature engineering Step (2). Here the researchers are guided by domain knowledge for creating new features from the raw data with the motivation that the derived features amplify the signal in the data which will then increase the accuracy of the machine learning models. This process is creative and involves considerable trial-and-error in conjunction with subsequent steps in the framework. 

A selection of new and original raw-data features is subsequently used to develop predictive machine learning models in Step (3). Multiple algorithms representing a wide variety are used to develop competing models. This is necessary since no one algorithm outperforms the rest across all possible domains and datasets\footnote{No Free Lunch Theorem \cite{wolpert1997no}}. The competing algorithms are tuned with respect to their hyperparameters to ensure that the model training is not misspecified, thus entailing an iterative approach. The models are robustly evaluated against a variety of metrics to identify the best candidate out of the competing models. In case of insufficient model accuracy, new features may be required, thus returning to Step (2).

All models are by definition a simplification of some phenomenon. However, as machine learning algorithms become more sophisticated, the induced models in turn become uninterpretable. The subsequent steps in the predictive analytics framework aim to expose the opaque mechanics of black-box models. This is achieved by simulating the behaviour of the primary model and in effect constructing a \textit{model-of-the-model} which is a simplification of the original while having human-understandable characteristics. These second-order models are commonly referred to as \textit{surrogate} or \textit{proxy} models.

In Step (4), the selected best-performing model is evaluated for interpretability. This step has a two-fold purpose. First, during the development stage, it serves as another means for achieving model validation by the researchers. However, more important for the current context, is that in this step the global-level mechanics of the model's behaviour can be revealed. In this step, analytics tools are used that depict which features are the key drivers of predicted outcomes, and how the various feature values positively or negatively impact the final predicted outcomes. The outputs of this step are communicated to key stakeholders so that trust in the black-box models can be fostered and the predictions relied upon. If the interpretation of the models raises concerns about the validity of the models and the underlying features, then the process returns to Step (2), and new features may need to be engineered.

In Step (5), we transition towards local-level explainability of the model by interrogating it on how precisely it has reasoned to classify a specific learner with a given outcome. The information in this step becomes particularly meaningful for two reasons. First, the academic advisors who manage interventions are able to gather information and gain a deeper understanding of which factors for each learner are the contributing factors to their predicted negative outcome. This step is also critical for legal compliance aspects where affected stakeholders have the right to understand how the autonomous decision-making systems have arrived at particular conclusions about them \cite{mathrani2021perspectives}.    

At this stage, the limits of what predictive modelling can offer are reached. Seeking to understand what exactly can be done to maximise better outcomes belongs within the scope of prescriptive analytics. 

\subsubsection*{Prescriptive Analytics Phase}

In prescriptive analytics, the aim is to leverage techniques like machine learning models to perform counterfactual or \textit{what-if} simulations. The end goal is to use the outputs of the simulations as a basis for automatically generating evidence-based feedback to learners concerning adjustments to their learning behaviour which may result in improved outcomes. The simulations entail specifying an alternative outcome for a given learner to the one that the predictive models have arrived at. In this study's context, we are seeking to define a new target outcome of a 'successful qualification completion' for an at-risk learner. The counterfactual modelling in Step (6) takes the predictive model and uses it to uncover a set of minimal changes that would need to take place in the learner's feature values for the desired outcome to be predicted for a specific learner. Naturally, for this to be useful counterfactual modelling needs to be performed using features upon which actionable steps can be taken. This means that a set of target features need to be selected like: engagement levels with the VLE, average assignment marks and the total number of on-time assignment submissions etc. The counterfactual modelling step is flexible. New models can be developed at this stage solely for this task whereby more actionable features are used, and immutable features like demographics (eg. learner age) are removed. 
Counterfactual outputs need to be configured to model changes only in the selected features upon which advisors desire to base their feedback. Additionally, each feature in this modelling process needs to be constrained within a range of values for them to be varied to generate feasible and realistic pathways to a desirable predicted outcome. Several possible pathways towards an alternative outcome for a learner can be generated and not all will be feasible. The unrealistic pathways need to be filtered and discarded.


Once the most realistic pathways have been selected, the next Step (7) entails two parts. The first requires that the engineered feature values be converted back into raw values so that they become interpretable and actionable for the learner. This step, therefore, has a dependency on Step (1) in the framework where the raw values reside and enable the engineered values can be converted back to them. The second part converts the numeric values into natural language text which can be dispensed to learners either directly via LA dashboards and emails or through conversations with academic advisors. Converting the counterfactual outputs into a human-readable format also makes the prescriptive feedback more usable for the academic advisors for the task of filtering and selecting the most suitable prescriptive feedback from all the candidate options. \hl{ This study proposes that the conversion is performed using ChatGPT and automated using available APIs. In order to prime the LLM to generate appropriate text, it is first necessary to devise an ideal template response and to extract the feature names and their values into a structured formal, such as JSON, from which prescriptive feedback text can be constructed. An example of ChatGPT prompt engineering, a response template and the JSON object are provided in the case study.}
Once the prescriptive feedback has been generated and the most suitable selected, then the intervention can be initiated - Step (8).    

\section{Methodology}\label{}

\subsection{Dataset}\label{}

\paragraph{ Setting: } The data was acquired from an Australasian HEI. The data was sourced from the institution's Student Management System (SMS) and the Virtual Learning Environment (VLE) - Moodle. The dataset consisted of undergraduate students who commenced studies from 2018 through 2022 and who either completed or abandoned their studies during this period. The two categories of outcome represented an evenly balanced target variable where the total number of completed learners accounted for 52\% (3693), and those who abandoned their studies comprised 48\% (3415). 

\paragraph{Predictive problem formulation: } The target prediction variable was programme/qualification completion. Given the nature of the underlying data, each student represents a data point at a given point in time denoted by the academic year. Most students are enrolled over multiple academic years of their programme, and therefore the dataset represents students as snapshots across each of their years of enrollment. For example, if a student was enrolled in a three-year bachelor's qualification with ultimately a successful completion, then the student would be represented in the dataset with three data points and each would be designated the target variable of 'completed'. As a result, the total dataset consisted of 14918 data points. The learners who completed their programme of study comprised 72\% (10736) of all data points since they enrolled in multiple academic years, while non-completion learners comprised 28\% (4182), thus making the final dataset relatively unbalanced. 

Given the nature of the dataset with a learner's educational journey represented by multiple data points, the predictive problem, in this case, was both \textit{formative} and \textit{summative}. In the formative approach to the prediction of learner outcomes, students' outcomes are considered at various checkpoints of their studies of their journey towards completion.  However, in the summative prediction, the learner outcomes are predicted at the end of the qualification or semester when course-level predictive models are used. 

\paragraph{Features: } The features used to describe each learner at a particular point in time can be grouped into four broad categories. These categories are learner academic performance and learning behaviours, as well as the more immutable learner characteristics data. Additionally, these features were augmented with the characteristics of the programme into which they enrolled. 

Once the data was prepared (Step 1), raw feature values were engineered to extract new derivative features to create more descriptive feature values - Step (2). In order to draw out more value from the raw behavioural and academic performance features such as a learner's average grade or the number of online learning resources they accessed, these were transformed and relativised to the learner's cohort. This conversion made the feature values more generic, contextually meaningful and comparable across different cohorts who study different courses/programmes that have different means and spreads across various features. To achieve this, each course-specific feature was transformed into a normalised value using z-score standardisation. This conversion converted the learner's absolute and mean values for a given feature into a form that captures the degree to which it deviates from the learner's cohort. The utility of this approach to creating course-agnostic features was underscored by \citet{ramaswami2022developing}. The z-score calculation Formula \ref{zscore} denotes $x$ as the value being converted, $\mu$ and $\sigma$ represent the mean and the standard deviation of all the values of $x$: 

\begin{ceqn}\label{zscore}
\begin{align}
    z-score = \frac{x-\mu}{\sigma} 
\end{align}
\end{ceqn}

In essence, the value of the z-score communicates how many standard deviations a given value is from the mean. For a z-score equal to 0, it implies that the value is directly on the mean. A positive z-score signifies that the raw $x$ value is above the mean and the opposite holds for negative z-scores. The z-score values possess a greater descriptive power leading to more accurate machine learning models, but this comes with the cost of lower interpretability. All features used in the study are summarised and described in Table \ref{features}, with engineered features representing Step (2) being identified with $\bigstar$.

\begin{table}[hbt!] 
\caption{Descriptions of features used for predictive and prescriptive modelling. $\bigstar$ denoting engineered features, while $\bigcirc$ describes features used only for prescriptive model generation and $\oplus$ denotes features used for both prescriptive model generation and prescriptive feedback generation. }\label{features}
\fontsize{7}{7.2}\selectfont
\centering
\begin{tabular} {p{0.17\textwidth}@{\extracolsep{\fill}} p{0.25\textwidth}p{0.3\textwidth} >{\centering\arraybackslash}p{0.1\textwidth} >{\centering\arraybackslash}p{0.07\textwidth}}
\toprule
Category     & Feature Name      & Description   & Predictive Modelling & Prescriptive Modelling  \\
\midrule
LEARNER CHARACTERISTICS         & Basis For Admission Description                       & Eg. NZ Entrance, NCEA,  adult admission, discretionary entrance etc. & $\checkmark$                   &                         \\
                                                  & Has Previous Tertiary Study                           & yes/no                                                                                                                                                   & $\checkmark$                    &  \\
                                                  & Highest School Qualification Description              &  Eg. NZ University Entrance,  International Baccalaureate etc.                                                                                                                                                       & $\checkmark$                    & \\
                                                  & Current Full Time Status                              & full-time/part-time                                                                                                                                      & $\checkmark$                    & $\oplus$ \\
                                                  & Current Student Mode Numeric                          & On-campus or online/distance                                                                                                                                 & $\checkmark$                   & $\oplus$\\
                                                  & Current Prior Activity Description                    & What was the primary activity that the student was engaged in, in the previous year  & $\checkmark$&\\
                                                  & Age Description                                         & current student age  & $\checkmark$                   & $\bigcirc$                       \\
                                                  & Gender                                                   & male/female/other & $\checkmark$                   &                         \\
\cmidrule{2-5}
ACADEMIC PERFORMANCE DATA      & Grade Mark Mean $\bigstar$& Student’s mean grade for current   academic year courses  & $\checkmark$& $\oplus$ \\
& Grade Mark Max $\bigstar$& Student’s max grade for current academic year courses  & & $\bigcirc$  \\
 & Grade Mark Deviation From Class Mean $\bigstar$               & Student’s Z-score calculation with respect to the average cohort grade across all courses enrolled in & $\checkmark$                   & $\bigcirc$                       \\
                                                  & Papers Failed For Student Academic Year $\bigstar$           &  & $\checkmark$                   & $\oplus$                    \\
                                                  & Online Learning Has Passed Assessment Count Zscore $\bigstar$& The total number of assessments a student has passed represented as a relativised Z-score with respect to the student's current cohort values.         &                      & $\bigcirc$                       \\
                                                  & Total Qualification Percent Completed $\bigstar$               & Percent of the student's  qualification completed with respect to the number of courses required  &                      & $\oplus$                      \\
                                                  & Online Learning Submitted Assignment Zscore $\bigstar$        & The average assignment mark for a student  represented as a relativised Z-score with respect to the student's current cohort average.              & $\checkmark$                   & $\oplus$                      \\
\cmidrule{2-5}
LEARNER  BEHAVIOUR DATA         & Papers Withdrawn For Student Academic Year  $\bigstar$       & Number of paper that a student has withdrawn from in a given year. & $\checkmark$                   & $\oplus$                      \\
                                                  & Online Learning Pages Viewed Count Zscore $\bigstar$         & The total amount of  VLE content accessed by a student represented as a relativised Z-score with respect to the student's current cohort totals
                                                    & $\checkmark$                   & $\oplus$                      \\
                                                  & Online Learning Quiz Taken Count Zscore  $\bigstar$          & The total number of quizzes taken by a student represented as a relativised Z-score with respect to the student's current cohort totals                  & $\checkmark$                   & $\oplus$                      \\
                                                  & Online Learning Forum Post Created Count Zscore  $\bigstar$ & The total number of discussion forum posts created by a student represented as a relativised Z-score with respect to the student's current cohort totals 
                                                               & $\checkmark$                   & $\oplus$                      \\
                                                  & Online Learning Forum Post Read Count Zscore $\bigstar$     & The total number of discussion forum posts read by a student represented as a relativised Z-score with respect to the student's current cohort totals               & $\checkmark$                   & $\oplus$                      \\
                                                  & Online Learning On Time Submission Count Zscore $\bigstar$  & The total number of on-time assignment submissions by a student represented as a relativised Z-score with respect to the student's current cohort totals     & $\checkmark$                   & $\oplus$                      \\
\cmidrule{2-5}
PROGRAMME CHARACTERISTICS DATA & Programme Title        & Name of qualification, eg. Bachelor of Arts    & $\checkmark$                   &        \\
                                                  & Programme Credits Required                             & Eg. 60, 120, 360, 480. & $\checkmark$& $\oplus$ \\
& & & & \\

\bottomrule                                                  
\end{tabular}
\end{table}

\subsection{Predictive Machine Learning}\label{}

\paragraph{ Algorithms: } A wide variety of algorithms from a broad range of machine learning families of techniques were used for the experiments to generate candidate models as part of Step (3). These consisted of Scikit-learn \cite{ScikitLearn2022} Python implementations of Random Forest (RF) \cite{breiman2001random}, K-Nearest Neighbour Regression (kNN) \cite{cover1967nearest}, Naive Bayes (NB), Support Vector Machines (SVM), Gradient Boosting (GB), Logistic Regression (LR),  Decision Tree (DT) and CatBoost (CB) \cite{catboost}\footnote{ This is a non-scikit-learn catboost 1.0.6 implementation found here: https://pypi.org/project/catboost/  }. Two baseline models were used for demonstrating the predictive value of the candidate models, namely, the stratified random guessing model (Baseline 1) as well as the mode (Baseline 2). 

\paragraph{Data Preparation: } In cases where there were missing values and the underlying algorithms required the presence of all values, these were replaced with zero. For algorithms that required all values to be numeric, Binary Encoding of categorical values was used which produced more concise feature sets and reduced the likelihood of overfitting\footnote{CatBoost was the only algorithm in the suite of techniques used which did not require the imputation of missing values and the encoding of categorical values into numeric data types.}. 

\paragraph{Training Approach and Hyperparameter Tuning: } Training was performed in such a way as to prevent data leakage from occurring which would compromise the validity of the predictive modelling results. Given that each learner is potentially represented by several data points, the experimental design involved applying train/test splits in a manner that ensured all data points from a given learner were either in the training or the test set. 

All algorithms possessing consequential hyperparameters were first tuned using Random Grid Search in conjunction with a 5-fold cross-validation approach. The best performing hyperparameters were then used on a separate 10-fold cross-validation process to collect the estimated generalisability scores. The range of hyperparameters for each algorithm and the best performing values can be seen in Table \ref{tuning}. The final predictive models were evaluated using the overall F1-measure due to the unbalanced target variable. F1-measure calculates the harmonic mean of Precision and Recall values. However, for completion, the Area Under the Curve (AUC), Accuracy, Recall and Precision metrics are also reported separately as an average value across all test folds, together with their dispersion in the form of the standard deviation.

\begin{table}[hbt]
\caption{Optimal hyperparameters resulting from random search tuning. }\label{tuning}
\begin{tabular}{>{\raggedleft}p{0.25\textwidth}>{\raggedright\arraybackslash}p{0.7\textwidth}}

\hline 
Algorithms  & Optimised hyperparameter values  \\ 
\hline 
SVM & kernel=rbf, gamma=0.0001, C=1000 \\
Decision Tree & max\_depth=10, criterion=entropy \\
Logistic Regression & solver=liblinear, penalty=l2, C=336 \\
CatBoost & depth=7, iterations=150, learning\_rate=0.07 \\
kNN & weights=distance, p=1, n\_neighbors=5, metric=chebyshev, leaf\_size=20 \\
Random Forest & n\_estimators=200, min\_samples\_split=2, max\_features=auto, max\_depth=9 \\
Gradient Boosting & subsample=0.85, n\_estimators=500, min\_samples\_split=0.47, min\_samples\_leaf=0.115, max\_features=sqrt, max\_depth=5, learning\_rate=0.2, criterion=mae \\ 
\hline 
\end{tabular}
\end{table}

\subsection{Explainable AI Tools}\label{}

SHAP (SHapley Additive exPlanation) which is based on Shapley values \cite{shapley1953quota} drawn from game theory literature,  was used to generate \textit{global} interpretability defined in Step (4), as well as \textit{local} model explainability required for Step (5), addressing both the "how" and "why" questions around model behaviour respectively. 
In addition, the Anchors tool was also used in Step (5) alongside SHAP. Both techniques approach the generation of new \textit{surrogate models} approximating the behaviour of the original "black-box" models differently and offer complementary insights. The advantage of SHAP lies in its detailed quantification of effects that each feature and their values exert on the final prediction, while Anchors reduces a black-box model into a human-readable set of predicates resembling a degenerate decision tree.

For the prescriptive analytics phase, the Diverse Counterfactual Explanations (DiCE) \cite{mothilal2020dice} technique was used in Step (6) to simulate \textit{what-if} scenarios and generate candidate prescriptive feedback suggestions. 
DiCE can generate counterfactuals for many machine learning models, approaching the task as an optimisation problem. However, counterfactual modelling is a challenge and needs to be configured in such a way that, firstly, the features used are relevant for a learner and represent learning behaviours that can be adjusted, and secondly, the suggested action must be feasible and practical \cite{poyiadzi2020face}. To that end, a new underlying machine learning model was generated using an alternative set of features comprising predominately mutable features which would support the development of actionable feedback. These features can be seen in Table \ref{features} under the Prescriptive Modelling column. However, the counterfactual modelling was further configured to exclusively rely on features upon which prescriptive advice was to be based. Table \ref{features} therefore shows features that were used for creating a new prescriptive model as denoted by $\bigcirc$, and features that were used for both modelling and for generating counterfactuals $\oplus$. Secondly, the features used for counterfactual modelling have been constrained to fall within feasible and realistic ranges. Finally, once the candidate counterfactuals were generated, these were then converted into human-readable text as indicated in Step (7).

\subsection{Framework Evaluation}\label{}

The quantitative component of the framework concerning predictive analytics is evaluated through empirical experiments that demonstrate the accuracy of the generated models, which also validate the efficacy of the engineered features (Steps 2 - 3).

The remaining parts of the framework are qualitative. Steps 4 and 5 are evaluated by assessing the suitability of the tools to visually convey the mechanics of the underlying model and to establish its reasonableness.  
A case study using two hypothetical students is used to evaluate in detail the model behaviour in explaining its predictions for Step 5.  
The same case study examples are then carried through to the prescriptive analytics component (Steps 6 - 7), where several prescriptive feedback suggestions are generated for each student, and subsequently converted into human-readable text for assessment of their feasibility.

\section{Results}\label{}

The efficacy of the candidate predictive models to identify learners at risk of programme/qualification non-completion is evaluated initially, and the best algorithm for this dataset is identified (Step 3). The characteristics of the best-performing predictive model are next examined to interpret its mechanics (Step 4). A case study involving two hypothetical students with non-completion predictions is then conducted to demonstrate how model explainability can be leveraged to interrogate the models about their reasoning for arriving at given predictions (Step 5) - two approaches to achieve this are presented. The application of prescriptive analytics is subsequently demonstrated using the same two hypothetical students for illustrative purposes. The capability of counterfactual modelling (Step 6) to derive potential prescriptive feedback for learners is then shown. Finally, the conversion of the prescriptive feedback into human-readable text suitable for learners (and advisors) is demonstrated (Step 7).   

\subsection{Predictive Model}\label{}

Table \ref{accuracies} summarises the generalisation performances of all candidate models across several evaluation measures. The results indicate that overall, the models have achieved a high level of efficacy for predicting whether a learner would eventually complete or abandon their studies. The preferred F1-measure on the unbalanced dataset indicates that the values range from ~92\% to ~95\% and represent a significant improvement over the baseline models. The table highlights the importance of reporting the accuracies of baseline models so that the genuine value of the predictive models with respect to various forms of random guessing can be quantified. On average, the best-performing algorithm for this dataset is CatBoost, which is slightly better than Random Forest. Some of the performance advantages of CB can arguably be attributed to its inherent ability to handle both missing values and categorical data directly in contrast to the other algorithms in these experiments.     

\begin{table}[hbt]
\caption{Predictive accuracies of all candidate models listed in a descending order of estimated generalisability using the F1-measure. }\label{accuracies}
\centering
\begin{tabular}{rlllll}
\hline
Algorithm & F1-measure & Accuracy                  & AUC                                              & Recall                    & Precision                 \\
\hline
CatBoost	  & 94.5	$\pm$0.6   & 92.0	$\pm$0.7	  & 88.3	$\pm$0.6	  & 96.7	$\pm$0.7	  & 92.5	$\pm$0.7\\
Random Forest	  & 93.9	$\pm$0.6   & 90.9	$\pm$0.9	  & 85.6	$\pm$1.2	  & 97.7	$\pm$0.5	  & 90.4	$\pm$1.0\\
SVM	  & 93.8	$\pm$0.6   & 90.7	$\pm$0.9	  & 85.8	$\pm$1.2	  & 97.1	$\pm$0.7	  & 90.7	$\pm$1.1\\
Logistic Regression	  & 93.5	$\pm$0.8   & 90.3	$\pm$1.1	  & 85.5	$\pm$1.4	  & 96.5	$\pm$0.8	  & 90.7	$\pm$1.1\\
Decision Tree	  & 93.0	$\pm$0.6   & 89.6	$\pm$0.8	  & 84.1	$\pm$1.3	  & 96.6	$\pm$0.4	  & 89.7	$\pm$1.1\\
Gradient Boosting	  & 92.9	$\pm$0.5   & 89.4	$\pm$0.7	  & 84.4	$\pm$0.8	  & 95.9	$\pm$0.8	  & 90.0	$\pm$0.7\\
kNN	  & 92.9	$\pm$0.7   & 89.3	$\pm$0.9	  & 83.5	$\pm$1.2	  & 96.8	$\pm$0.5	  & 89.3	$\pm$1.1\\
Naive Bayes	  & 91.5	$\pm$0.6   & 87.6	$\pm$0.8	  & 83.2	$\pm$0.9	  & 93.2	$\pm$0.7	  & 89.9	$\pm$0.8\\
Baseline 2 	  & 83.7	$\pm$1.0   & 71.9	$\pm$1.5	  & 50.0	$\pm$0.0	  & 100.0	$\pm$0.0      &	71.9	$\pm$1.5\\
Baseline 1	  & 72.0	$\pm$0.9   & 59.6	$\pm$1.0	  & 49.6	$\pm$1.0	  & 72.4	$\pm$1.7	  & 71.7	$\pm$1.0\\
\hline
\end{tabular}
\end{table}

\subsection{Model Interpretability}\label{}

\begin{figure}[hbt]
	\centering
		\includegraphics[scale=0.45]{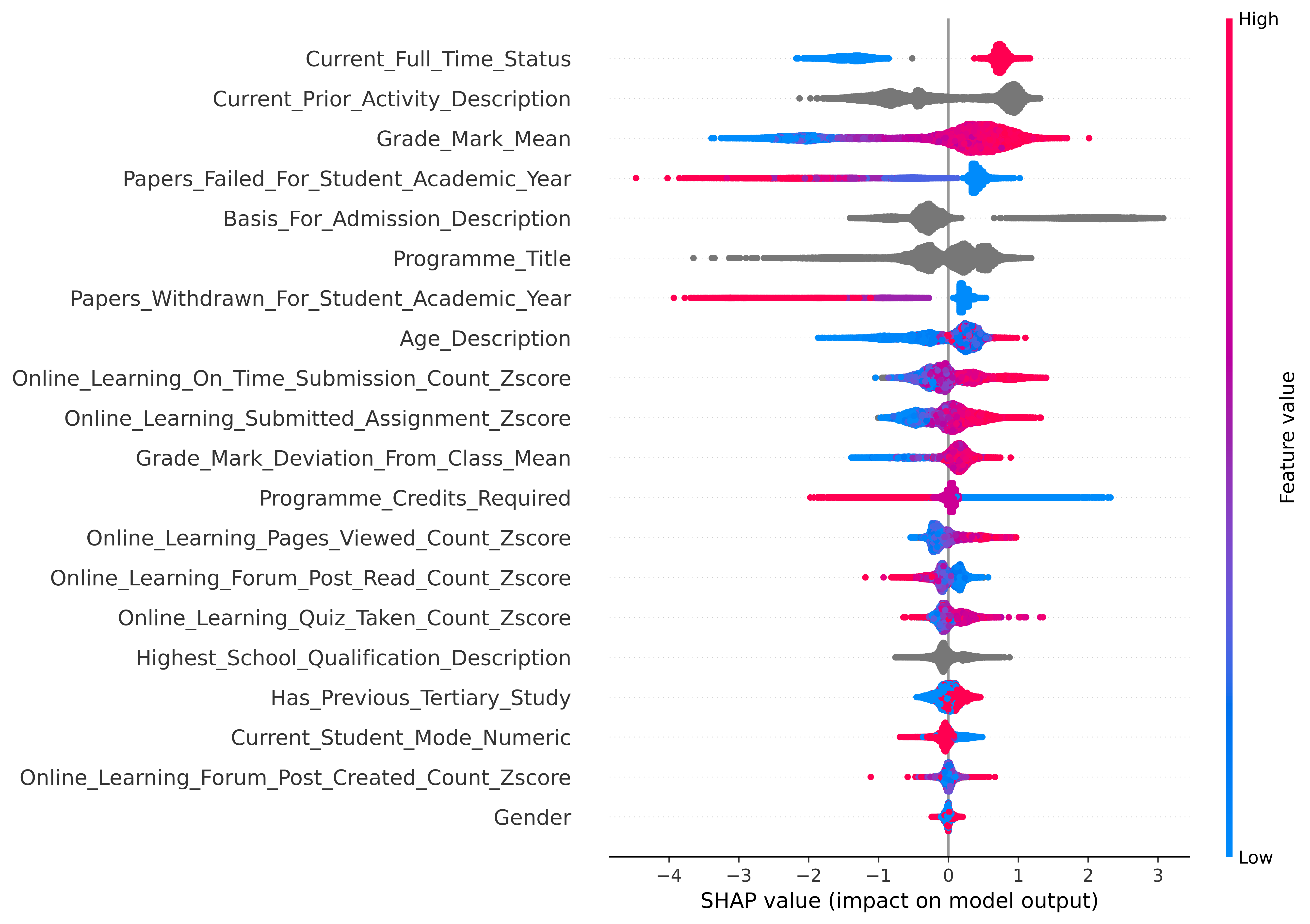}
	  \caption{SHAP summary plot showing the high-level behaviour of the CatBoost model. Most impactful features are shown from top to bottom. }\label{shapsummary}
\end{figure}

Having determined the efficacy of the model to identify students at risk of programme non-completion, the global-level interpretability of the best-performing CatBoost model is next examined.   
Figure \ref{shapsummary} depicts SHAP's perspective of the model's dynamics. Two key components are shown. The first lists features in their order of importance, from highest to lowest in terms of the impact they exert on the eventual prediction. It can be seen that the learners' current full-time status, their prior activity with respect to the current academic year, the mean grade mark, together with the number of failed papers are the most impactful features.

The second component in the figure offers an additional dimension concerning the interpretability of the model. Here we observe how an increase or decrease in feature values affects the final prediction. The colour gradients represent increasing (red) and decreasing (blue) feature values, while grey represents categorical values. The x-axis depicts SHAP values. Data points with a positive SHAP value (appearing to the right of the vertical zero line) have a positive impact on the predictions, in other words, they contribute towards driving the prediction towards programme completion predictions. Conversely, the points with a negative SHAP value (to the left of the vertical zero line) influence the prediction towards programme non-completion. The extent of the points from the vertical line signifies the magnitude of the effect that they contribute to the final prediction.

In the figure, it can be observed that the full-time study status (value 1) has a positive effect on completion predictions, while part-time (value 0), has the opposite. Learners' mean grade has an unsurprising effect; however, a nuanced interpretation can be extracted from the plot where one can deduce that lower grade averages have more of a negative predictive outcome than high grade-averages have on positive outcomes. A similar interpretation can be made regarding the number of failed papers. A large number of failed papers has more of a negative predictive effect than having no failed papers has on positive predictive outcomes. 

In general, elevated assignment scores and submission counts are predictive of positive outcomes. Learner engagement with the VLE conveys a more mixed picture. VLE pages viewed and the number of online quizzes taken have a positive effect; however, forum post creation counts are ambivalent, while the reverse holds for the number of forum posts read. As the number of qualification credits increases, the effect is stronger for negative outcomes, while the reverse holds for the learner's age. Possessing previous studies (value 1) indeed has a more positive effect than otherwise (value 0), while the on-campus study mode (value 1) is associated with more negative qualification outcomes than studying online (value 0). 

By considering both the feature importance ranks and how the feature values affect the final prediction, it is possible to validate the model against an expected behaviour and communicate its mechanics in a simplified form to all stakeholders. The behaviour of the examined model confirms that it is reasonable and thus valid.

\subsection{Model Explainability}\label{}

Having achieved interpretation and validation of the predictive model, the next step is to examine the model behaviour at an individual (or local) prediction level. SHAP as well as Anchors\footnote{An alternative technology for this is LIME which has some additional advantages.} are used in this step. To demonstrate this, two hypothetical students are used - Student A and Student B. Both students have been predicted by the CatBoost model with non-completion outcomes, with the probability of 97\% and 90\% respectively.

\begin{figure}[hbt]
	\centering
\subfloat(a){%
  \includegraphics[clip,width=0.85\columnwidth]{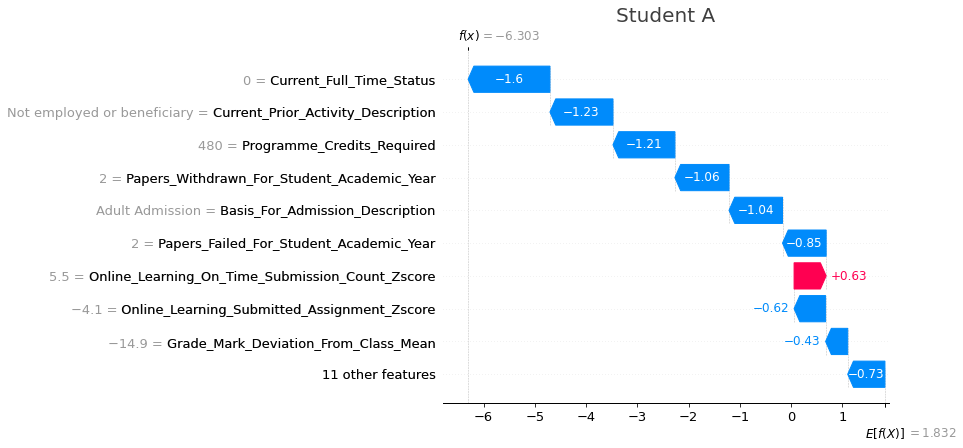}%
}
\subfloat(b){%
  \includegraphics[clip,width=0.85\columnwidth]{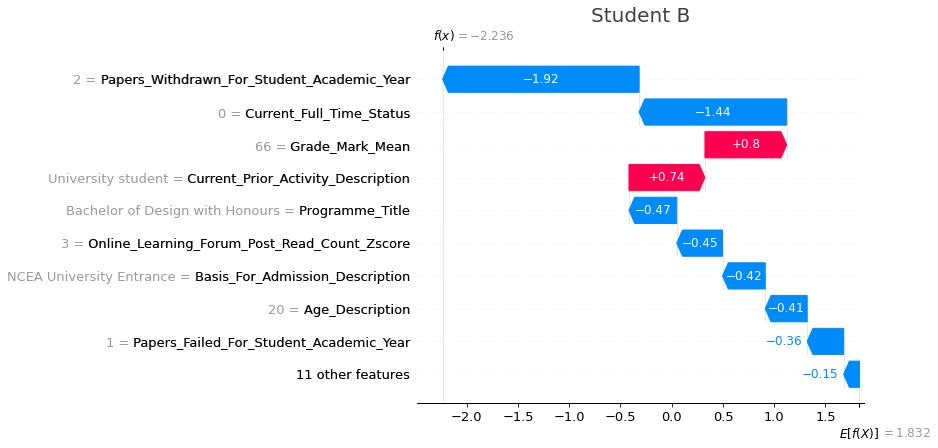}%
}
\caption{SHAP force plot for Student A and B depicting the effects that their feature values have on the final prediction outcome. }\label{shapwaterfall}
\end{figure}

Figure \ref{shapwaterfall}(a) shows the top nine features and their values for Student A on the y-axis, rank-ordered by influence on the final prediction. Informally, the figure can be viewed and interpreted as a tug-of-war. The mid-line represented with the value of 1.823 is the expected or the average SHAP value of all the predicted data points. A final SHAP value to the left of this line represents a non-completion prediction and alternatively, the values on the right side denote positive outcome predictions. Blue bars represent the forcing effects towards negative predictions, while red the opposite. The size of the bars represents the magnitude of the forcing of the corresponding features and their values. These graphs are best interpreted from the bottom up. The topmost feature represents the final SHAP value that includes all feature contributions.

Figure \ref{shapwaterfall}(a) shows that the least significant 11 features collectively have an influence tending towards negative outcomes to the left of the 1.823 value mid-line. This increases with each of the features except for the student's learning behaviour for on-time assignment submissions, which has a positive effect.

Model reasoning is depicted in Figure \ref{shapwaterfall}(b) for Student B. Similar patterns are observed with the exception that this student is a returning learner (prior activity is 'university student') and that the student's mean grade is 66\%, which have a strong positive influence on the final prediction. However, in totality, the majority of the features are forcing the prediction towards negative outcomes, and this is where they ultimately settle. The utility of the SHAP tool to visually explain its reasoning to academic advisors pre-intervention as well as to other relevant stakeholders is demonstrable through these examples. 

\begin{figure}[hbt]
	\centering
\subfloat(a){%
  \includegraphics[clip,width=0.9\columnwidth]{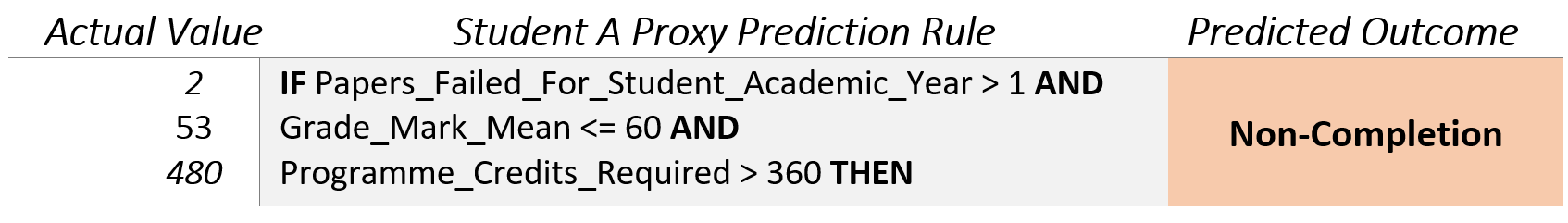}%
}
\subfloat(b){%
  \includegraphics[clip,width=0.9\columnwidth]{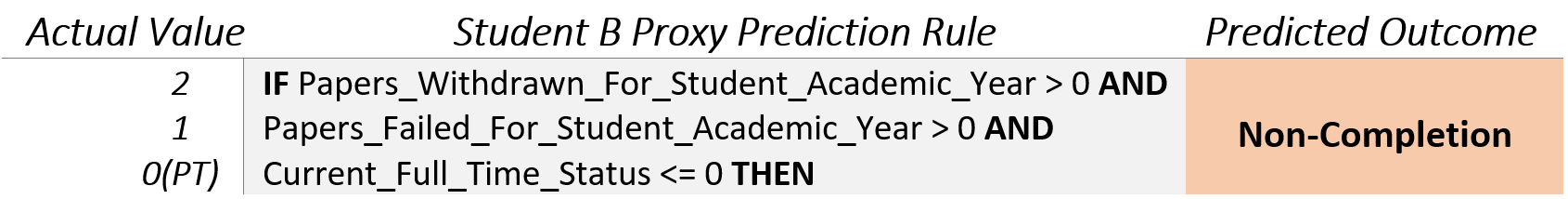}%
}
\caption{Proxy model explanation of predictions for Student A and B.}\label{proxy}
\end{figure}

While SHAP provides a detailed visual perspective into the model mechanics, some cccasions will require that a more succinct and simplified explanation of a prediction is communicated.
This can be achieved using Anchors. Figures \ref{proxy}(a) and \ref{proxy}(b) show how the complex predictive model can be reduced to the most essential and explained in a more simplified manner for both Student A and  B respectively. The figures show that the prediction model has been re-cast as a rule-based decision tree consisting of only three conditions, which result in a non-completion prediction if they all hold true. Student A and B's actual values for the three features are shown in the first column with the conditional statements and their thresholds for each feature shown in the second column.

In both cases, the Anchor surrogate model has identified and used the number of papers failed as one of the reasons for a negative prediction. For Student A, the fact that the student has a low average grade of 53\% while undertaking a commitment to a programme that is higher than the standard bachelor's qualification (360 credits), has been used as further conditions for the classification as non-completion. In the case of Student B, the fact that the student has already withdrawn from papers in the current academic year and is presently a part-time student, has created conditions for a non-completion predicted outcome. These forms of model simplifications serve as effective and suitable tools for communicating to affected learners how exactly they have come to be identified as being at-risk, thus meeting the requirements of transparency and responsible use of predictive analytics.

The generation of surrogate models as demonstrated above already hints at possible prescriptive suggestions which can be constructed from them.  Indeed, there is a potential to do this, however, a more data-driven and robust method ought to be pursued which considers the interaction of the features and their effects in a more principled approach. This is where prescriptive analytics tools make their contribution.

\subsection{Prescriptive Modelling}\label{}

\begin{figure}[hbt]
	\centering
\subfloat(a){%
  \includegraphics[clip,width=0.9\columnwidth]{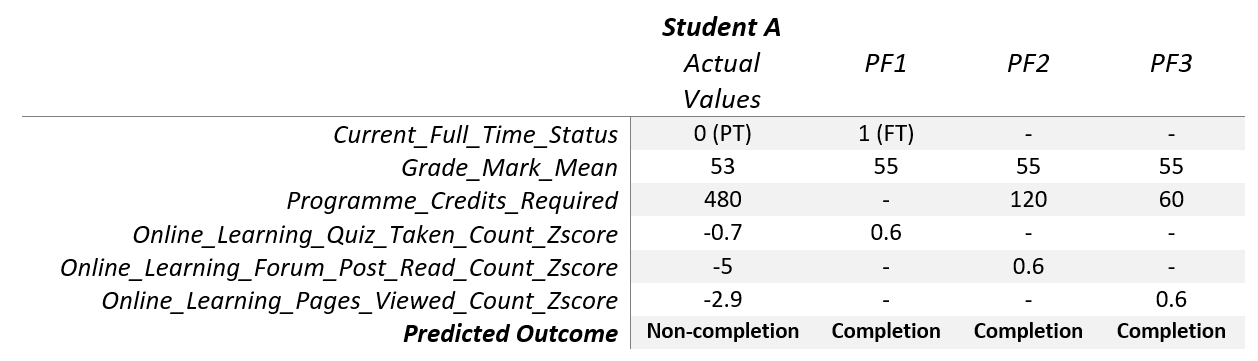}%
}
\subfloat(b){%
  \includegraphics[clip,width=0.9\columnwidth]{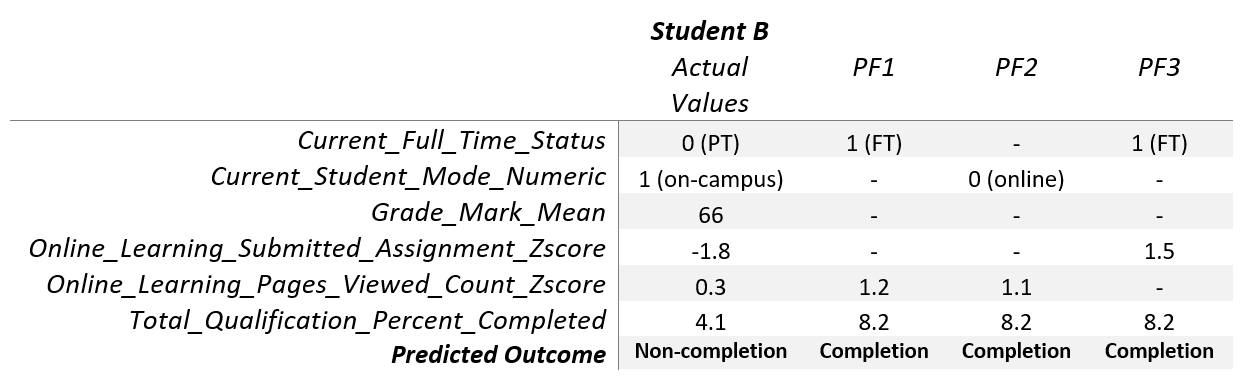}%
}
\caption{Depiction of three sets of Prescriptive Feedback (PF) options generated using counterfactuals for Student A and B. }\label{cfs}
\end{figure}

Counterfactual modelling is used in this step to generate a set of possible pathways for a specific learner that would lead them to a positive outcome prediction. In more precise terms, here we are looking at several possible sets of minimal adjustments to selected feature values which would result in an alternate outcome for a student who is predicted to be on track for non-completion. This type of \textit{what-if} modelling is demonstrated for Student A and B in Figure \ref{cfs}.

Figure \ref{cfs} depicts both the most concise set of features and the smallest required adjustments in their values which would be needed for the selected students to toggle their predicted outcome to a successful completion. The first column lists the selected features by the counterfactual model, and the next column shows the actual values for each of the hypothetical students, followed by three sets of counterfactuals from which automated and data-driven Prescriptive Feedback (PF) advice can be generated. In each PF set, only three feature values have been varied. The dash represents no required changes to the actual values.

In Figure \ref{cfs}(a), it can be seen that modest increases to the grade average have been identified as a pathway to completion for Student A, together with a switch to a smaller programme of study, as well as a mixture of adjustments to the online learning engagement behaviours. In the case of Student B, Figure \ref{cfs}(b) shows multiple pathways towards completion if the study mode is changed to full-time as well as to online mode on another occasion. Pathways exist through some modifications in online learning behaviours, while no adjustment is needed to be made for the grade average of 66\%. Interestingly, the qualification percent completion feature has been identified in the case of this student as being potentially helpful. In all three PFs, it is observable that if the student succeeds in completing 8.2\% of their programme, up from the current 4.1\%, then this also suggests that positive outcomes become more likely.

From this, it becomes immediately apparent that z-score values and percentages of completion carry with them very little meaning and actionable potential on behalf of the target learners as well as for the academic advisors. It is for this reason, that the engineered features need to be converted back into original raw values which will then make them practical. 

\subsection{Remedial Advice Generation}\label{}

The final step in the proposed framework performs two types of conversions. The first converts the engineered features chosen by the counterfactuals into raw values\hl{ so that they are meaningful to the learners. } The second step invokes ChatGPT via an API and converts the candidate PFs into a natural language form which can then be dispensed to students. 

\begin{figure}[hbt!]
	\centering
\subfloat(a){%
  \includegraphics[clip,width=0.9\columnwidth]{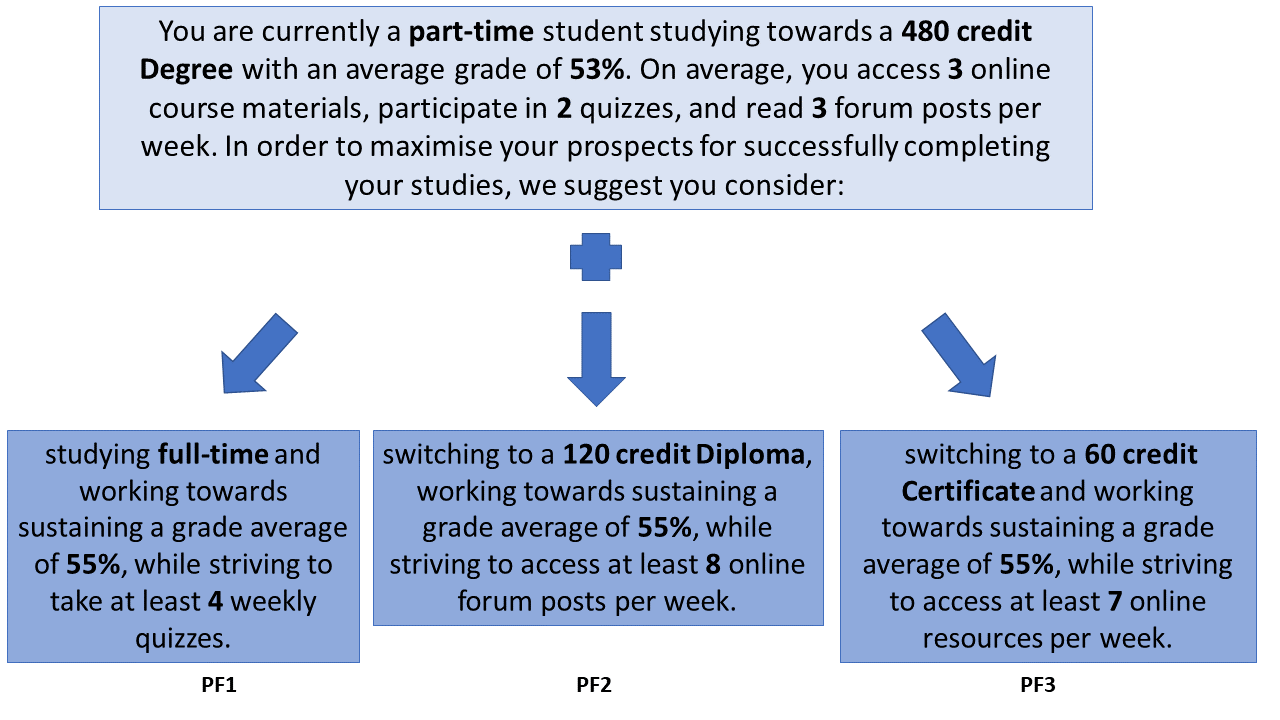}%
}
\subfloat(b){%
  \includegraphics[clip,width=0.9\columnwidth]{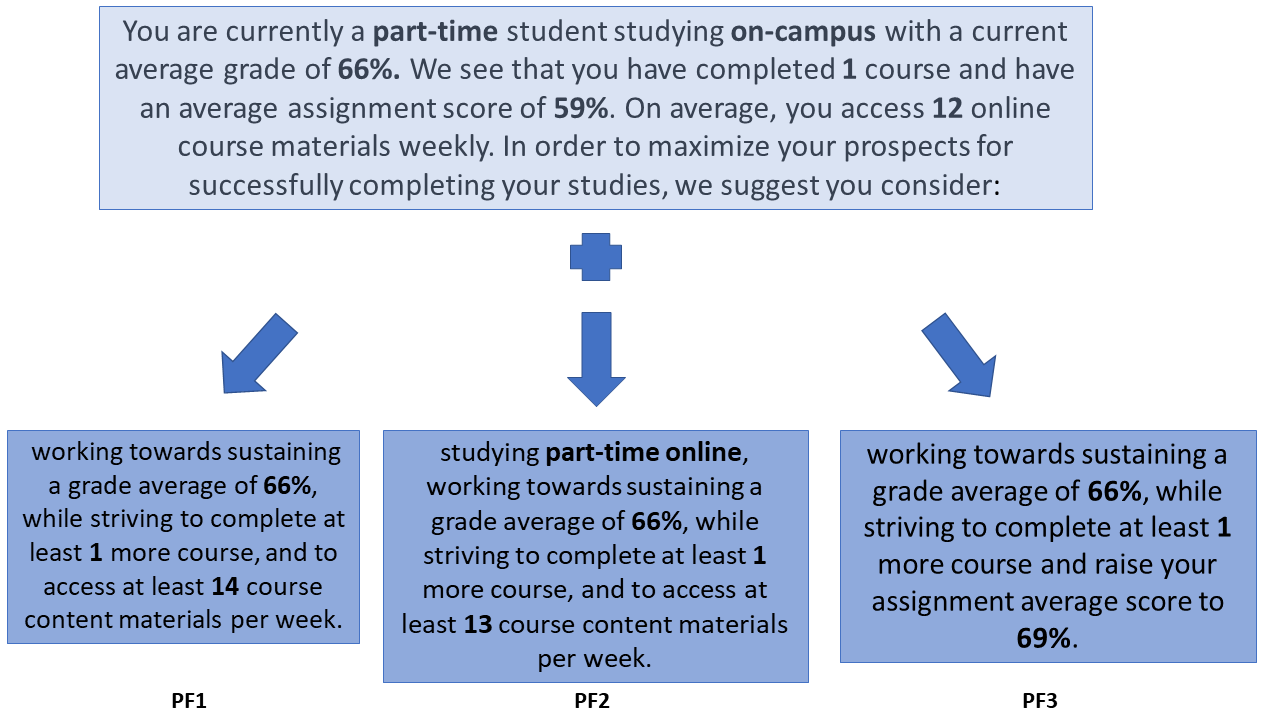}%
}
\caption{Depiction of the conversion of three sets of Prescriptive Feedback (PF) options into contextualised and human-readable student advice via ChatGPT, for Student A(a) and B(b), showing derived values from counterfactuals in bold. }\label{text_pfs}
\end{figure}

\hl{The result of converting PF counterfactuals into natural language can be seen in  Figure \mbox{\ref{text_pfs}} for Student A and B. The feedback consists of two parts. The first part contextualises the learners current status for key features that will be the basis for suggested remedial interventions and expresses their values. The second part suggests to the learner what changes to the specific features are likely to result in an alternative predicted outcome. The figure shows that both the z-scores and the programme completion percentages have been converted into meaningful values which are now actionable and measurable. This underscores the dependence between Step 7 and Step 1 in the proposed framework for the conversion of engineered values to raw values. }


\begin{figure}[hbt!]
	\centering
  \includegraphics[clip,width=0.9\columnwidth]{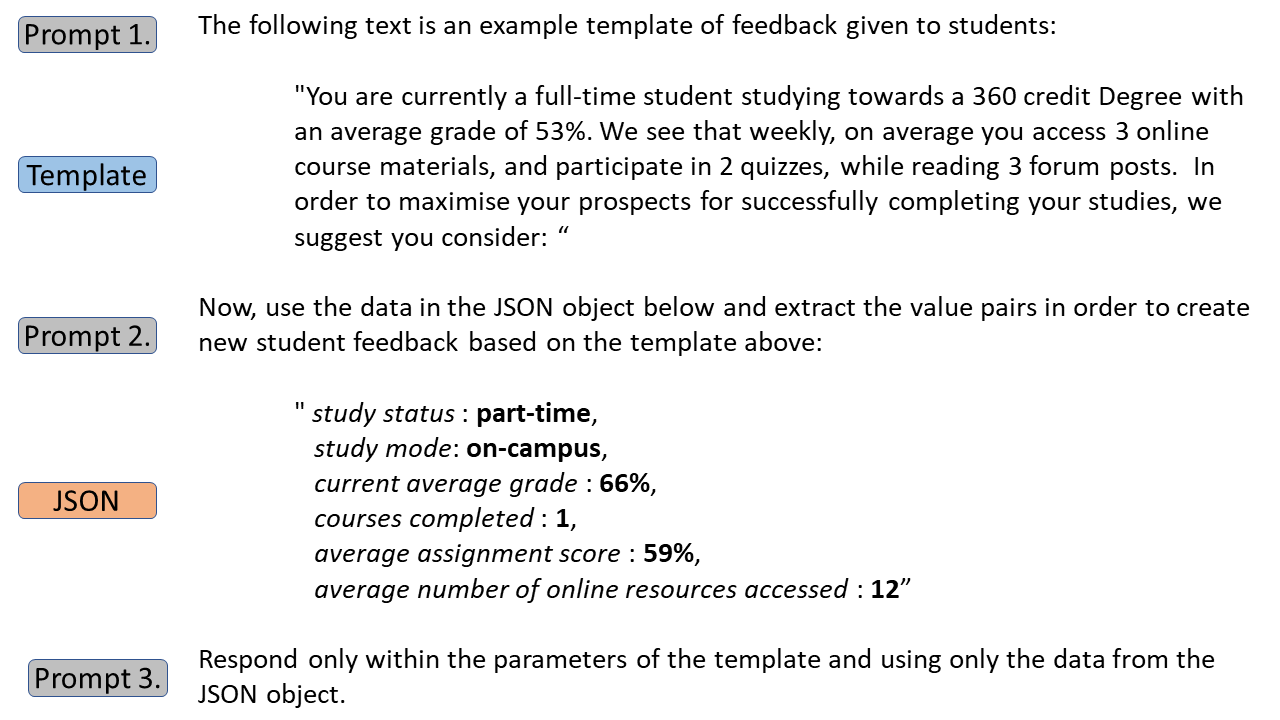}%

\caption{Example of the ChatGPT primer text for generating a learner current-status statement comprising the first part using Student B as an example. The primer text represents the contents of a single API call, containing the prompts, the response template and the data encapsulated in a JSON format. }\label{prompt1}
\end{figure}

\begin{figure}[hbt!]
	\centering
  \includegraphics[clip,width=0.9\columnwidth]{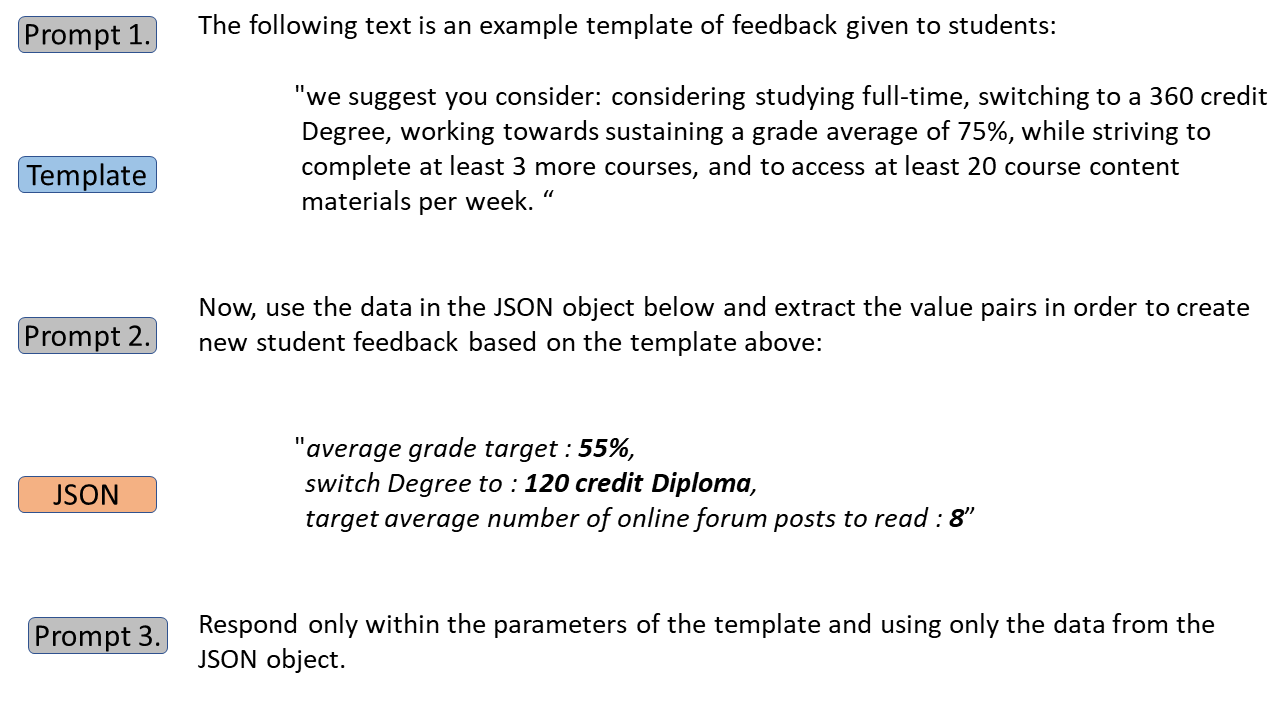}%

\caption{An example of the ChatGPT primer text for generating a second part of the prescriptive feedback text comprising the remedial suggestions. The primer text for Student B is given as an example. The primer text represents the contents of a single API call, containing the prompts, the response template and the data encapsulated in a JSON format. }\label{prompt2}
\end{figure}

\hl{The conversion of the counterfactuals into the natural language feedback is performed via ChatGPT and can be automated with existing APIs. Generating suitable prescriptive feedback using ChatGPT requires a systematic approach to prompt engineering that involves three parts. Firstly, it is necessary to provide ChatGPT with a well-defined prompt that outlines the purpose and scope of the feedback request. This prompt must be unambiguous in terms of the information that needs to be extracted and the context in which the feedback is being generated. Figures \mbox{ \ref{prompt1} and \ref{prompt2}} show how the prompts are devised and situated within the queries for the first and second parts of feedback generation respectively. }

\hl{Secondly, an example feedback template should be provided to serve as a reference and a primer for the output format that is expected from ChatGPT. This template acts as a guide for ChatGPT, helping to ensure that the feedback generated is consistent with the desired format. The prompts express clearly that the response must stay within the confines of the specified template.}

\hl{Finally, the actual data used to generate the prescriptive feedback must be provided. Here, the data is structured in a JSON object format. This format is conducive for automation via programming scripts and is understandable to ChatGPT for data extraction. The combination of these three steps ensures that ChatGPT can generate personalized and meaningful feedback based on the data provided without straying beyond requirements. Once the CFs have been converted into natural language, they can be more easily inspected and filtered by student support teams and the most suitable suggestions can be selected for interventions with learners as outlined in Step (8).   }
 

\section{Discussion}\label{}

The end goal of the proposed framework is the automated generation of evidence-based prescriptive feedback to learners who have been identified as being at-risk. The framework is generic and therefore the definition of at-risk is flexible and adaptable to any context. This study has demonstrated how this framework can be applied to learners who are at risk of programme non-completion.

In order to identify at-risk learners, the framework proposed how predictive analytics should be used to develop highly accurate predictive models, and how to use these models responsibly and with accountability. Using the models responsibly means exposing their internal mechanics and interpreting their behaviour to verify and validate them, and to be able to communicate this as plainly as possible to relevant stakeholders. This builds confidence and trust in the underlying black-box systems. Responsible use also means interrogating the models as to how they arrive at predictions for specific students which underscores accountability. The proposed framework outlines these steps and provides a clear roadmap in terms of which technologies and tools can be leveraged to support these tasks.

The proposed framework demonstrates how the leap from merely predicting outcomes to prescribing actions can be bridged for the first time in this domain using more advanced analytics. The danger in using counterfactuals in this domain is in assuming that causality is definitively established based on the fact that the prescriptive models have identified pathways to a positive outcome. This is not the case. The underlying prescriptive models rely on associations and cannot establish causality and this is a limitation.  

It must be emphasised that many latent variables cannot be captured that have a significant bearing on eventual learner outcomes. The features used in this study are largely proxies. We cannot capture variables describing a learner's true level of motivation, their sense of progress, and confidence which all have a bearing on eventual outcomes amongst many other variables. However, the possibility exists that if a learner is provided with actionable and achievable data-driven prescriptive feedback, and if it is followed and attained, the possibility exists that it may have a cascading effect on a learner's sense of achievement and thus on their level of motivation, progress and overall confidence, which may then lead to positive outcomes. 

Ultimately, the proposed framework demonstrates how multiple data-driven remedial advice options can be generated, \hl{ and translated into natural language via AI technologies like ChatGPT,} which can then be processed by academic advisors who are the human-in-the-middle, and they can then combine their experience and established theory of how to select the most suitable feedback suggestions for the target learners. Future work will involve evaluating the effectiveness of the proposed framework to improve student retention and qualification completion rates.

\section{Conclusion}\label{}

It has been a consistent feature of predictive Learning Analytics (LA) research targetting at-risk students, to focus exclusively on merely the predictive component. Predictive analytics is however much broader and it includes the unpacking of the internals of the predictive models' behaviour to stakeholders. It also encompasses responsible use of these automated systems which assist decision-making affecting humans. This includes the ability to interrogate the predictive models and seeks their reasoning as to how they have arrived at particular conclusions. eXplainable AI is a field that offers a suite of mature tools which enable this form of transparency that is largely absent in the current body of LA literature. 

What is more, predictions and their understanding while important, only address one part of the challenge in improving retention rates and increasing successful learner outcomes. Additional approaches are needed that can provide specific and tailored remedial advice to learners that are most likely to improve their outcomes. Prescriptive analytics tools support these aims and make any analytics endeavours more complete.

This work proposes a prescriptive analytics framework that demonstrates how both transparent predictive analytics can be achieved and combined with prescriptive analytics techniques. The study develops predictive models for identifying at-risk learners of programme non-completion. This work demonstrates through case studies how transparent and responsible predictive modelling can be augmented with prescriptive analytics to produce human-readable prescriptive feedback to those at risk \hl{via recent advances in AI using large language models}.

\urlstyle{same}



\end{document}